\documentclass[acmlarge]{acmart}

\AtBeginDocument{%
  }

\setcopyright{acmcopyright}
\copyrightyear{2023}
\acmYear{2023 (Preprint under review)}
\acmDOI{XXXXXXX.XXXXXXX}

\usepackage{listings}
\usepackage{subfig}
\usepackage{graphicx}
\begin{document}

\title{Mapping and Optimizing Communication in ROS 2-based Applications on Configurable System-on-Chip Platforms}

\author{Christian Lienen}
\email{christian.lienen@upb.de}
\orcid{1234-5678-9012}
\affiliation{%
  \institution{Paderborn University}
  \streetaddress{Warburger Str. 100}
  \city{Paderborn}
  \state{NRW}
  \country{Germany}
  \postcode{33129}
}

\author{Alexander Philipp Nowosad}
\email{anowosad@mail.uni-paderborn.de}
\affiliation{%
  \institution{Paderborn University}
  \streetaddress{Warburger Str. 100}
  \city{Paderborn}
  \state{NRW}
  \country{Germany}
  \postcode{33129}
}

\author{Marco Platzner}
\email{platzner@upb.de}
\affiliation{%
  \institution{Paderborn University}
  \streetaddress{Warburger Str. 100}
  \city{Paderborn}
  \state{NRW}
  \country{Germany}
  \postcode{33129}
}

\renewcommand{\shortauthors}{Lienen, Nowosad, and Platzner}

\begin{abstract}
The robot operating system is the de-facto standard for designing and implementing robotics applications. Several previous works deal with the integration of heterogeneous accelerators into ROS-based applications. One of these approaches is ReconROS, which enables nodes to be completely mapped to hardware. The follow-up work fpgaDDS extends ReconROS by an intra-FPGA data distribution service to process topic-based communication between nodes entirely in hardware. However, the application of this approach is strictly limited to communication between nodes implemented in hardware only.

This paper introduces gateways to close the gap between topic communication in hardware and software. Gateways aim to reduce data transfers between hardware and software by synchronizing a hardware-and software-mapped topic. As a result, data must be transferred only once compared to a separate data transmission for each subscribing hardware node in the baseline. Our measurements show significant speedups in multi-subscriber scenarios with large message sizes. From the conclusions of these measurements, we present a methodology for the communication mapping of ROS 2 computation graphs. In the evaluation, an autonomous driving real-world example benefits from the gateway and achieves a speedup of 1.4.
\end{abstract}



\maketitle

\section{Introduction}
\label{sec:Introduction}

\begin{figure}[!h]
	\center
    \includegraphics[width=\linewidth]{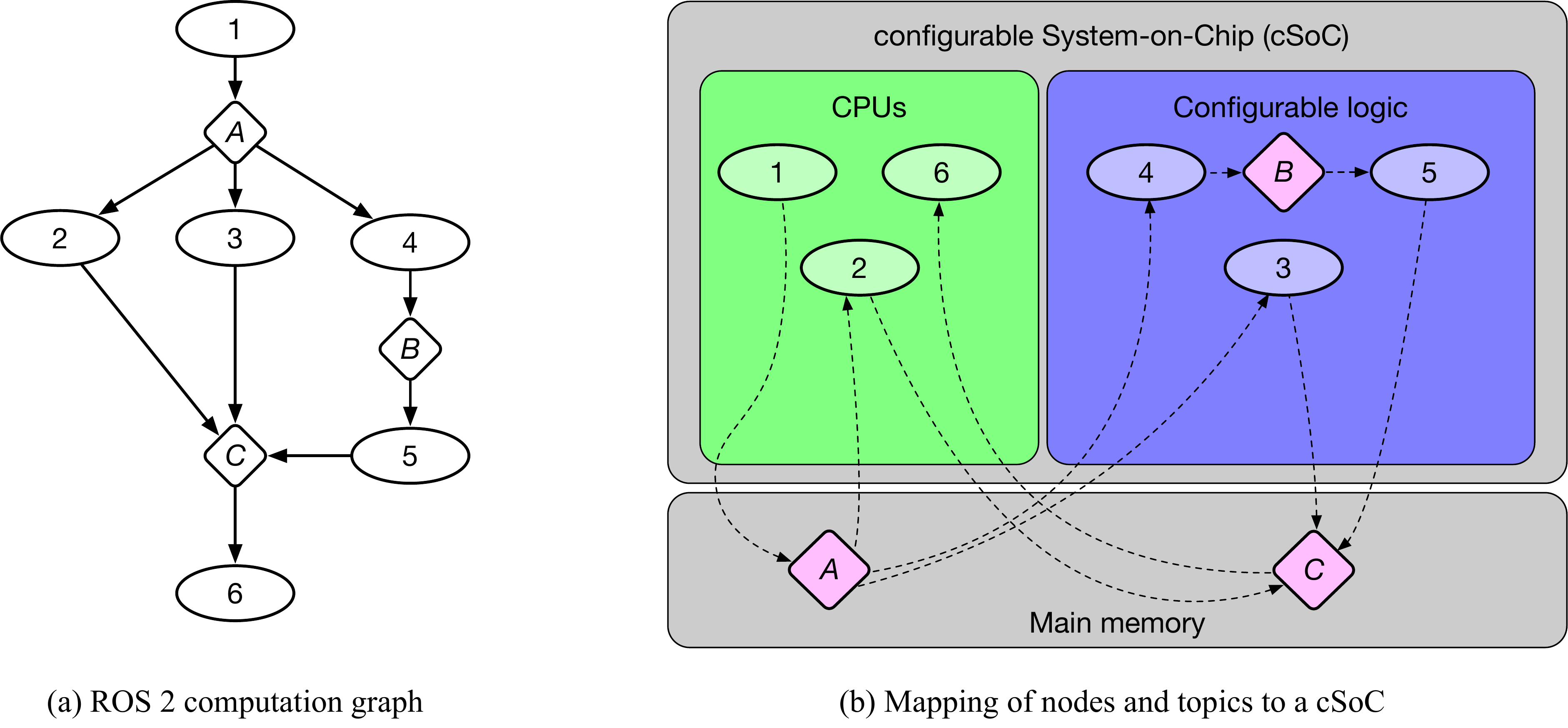}
    \caption{Example Application}
    \label{fig:problemdefinition}
\end{figure}

The de-facto software framework for developing robotics applications is the Robot Operating System, version 2 (ROS 2). ROS 2 decomposes an application into nodes, where each node is responsible for a part of the functionality of the overall application. ROS 2 provides different communication paradigms for exchanging data between the nodes, most importantly $n$-to-$m$ publish-subscribe communication via topics. %
ROS 2 applications are typically visualized by so-called computation graphs comprising nodes and their communication. Figure \ref{fig:problemdefinition}(a) shows an example computation graph with the six nodes 1-6 that communicate via the three topics {\em A, B}, and {\em C}. 

For execution, computational graphs are mapped to compute elements. Since ROS 2 is in essence a middleware wrapping communication functions in a layer termed data distribution service (DDS), the nodes can easily be distributed onto several networked computer systems. While ROS 2 nodes were mainly executed on CPUs in the past, modern state-of-the-art computing platforms for robotics offer heterogeneous execution units, such as multi-core CPUs, embedded GPUs, and FPGAs. Platform FPGAs are configurable system-on-chip (cSoC), combining, among other components, multiple CPUs and configurable logic. The flexibility and massive parallelism offered by the configurable logic are beneficial for many compute-intensive robotics functions, in particular, to improve latency and energy efficiency. 

In the last years, several approaches for enabling the computation of robotics functions on FPGAs have been presented. Most approaches map compute-intensive parts of ROS 2 nodes, so-called kernels, to the hardware. At runtime, the remaining parts of the ROS 2 nodes handle data transfer and results to and from this kernels~\cite {Yamashina2016}. An alternative and more flexible approach is ReconROS~ \cite{lienen2021design}, which allows for mapping complete ROS nodes to hardware. In systems where ROS nodes are mapped to both software and hardware, an efficient implementation of the communication functions is crucial for application performance. 

Figure~\ref{fig:problemdefinition}(b) 
presents an example mapping of the ROS application in Figure~\ref{fig:problemdefinition}(a) to a cSoC. ROS 2 nodes 1, 2, and 6 are mapped to software, and nodes 3, 4, and 5 are mapped to hardware. The standard mapping of topics is to software, which means the buffers holding messages are realized in the main memory external to the cSoC. Efficient DDS implementations such as Iceoryx~\cite{Iceoryx} are available to improve communication performance for shared memory architectures. However, communication to hardware-mapped nodes can challenge the memory interface of the configurable logic. For example, in the mapping of Figure~\ref{fig:problemdefinition}(b), messages from topic {\em A}, to which nodes 3 and 4 subscribe, are transferred two times to the configurable logic, reducing application performance. The DDS layer fpgaDDS~\cite{lienen2023fpgadds} has recently been presented for ReconROS. fpgaDDS maps topics completely to hardware if all nodes publishing to and subscribing from that topic are also mapped to hardware. Topic {\em B} in Figure~\ref{fig:problemdefinition}(b) exemplifies this case. However, one has to fall back to the standard mapping of topics to software or main memory, respectively, when at least one node publishing to and subscribing from a topic is mapped to software. 

In our work, we build on the open-source frameworks ReconROS and fpgaDDS to map ROS 2 applications to cSoCs. As a novelty, we introduce gateways and components that connect software-mapped and hardware-mapped topics in this paper. Using these gateways leads to efficient utilization of the memory interface of the configurable logic and, in turn, optimizes application performance. Finally, we present a methodology for mapping the topics of a ROS 2 computation graph to the execution platform utilizing software-mapped and hardware-mapped topics and gateways. 

The remainder of the paper is structured as follows: Section~\ref{sec:BackgroundRelatedWork} reports on related background, including ReconROS and fpgaDDS. In Section~\ref{sec:Gateway}, we elaborate on our gateway design, and in  Section~\ref{sec:Measurements} we characterize the performance of gateways. Based on our findings, we discuss the methodology for mapping topics in Section~\ref{sec:DesignMethodology}. In Section~\ref{sec:Evaluation}, we present an application case study before we conclude the paper in Section~\ref{sec:Conclusion}.

\section{Background}
\label{sec:BackgroundRelatedWork}

This section first provides an overview of concepts for hardware-accelerated ROS 2 applications and then introduces ReconROS and fpgaDDS.

\subsection{FPGA-based hardware acceleration of ROS applications}
\label{sec:BackgroundRelatedWork:HardwareAcceleration}

Several approaches for accelerating ROS 2-based applications on reconfigurable logic have been presented in recent years. Most of these approaches leverage remote procedure call patterns to offload runtime extensive code sections to the accelerator. The original ROS 2 nodes have to be modified to forward data to the acceleration kernels and receive their results. Examples can be found in~\cite{Yamashina2016,reconfros} or in the industry product Xilinx KRIA~\cite{mayoral2021kria}, for example.

Optimizing communication between nodes was also subject of research. Sugata et al.~\cite{Sugata2017} split communication into control and data communication and accelerate the latter by handling it in hardware. In follow-up work~\cite{8823798}, they simplify the development process by interpreting ROS messages during the hardware generation tool flow.
Communication between several acceleration kernels is optimized in~\cite{mayoral2022robotcore} by using streaming queues to achieve higher bandwidth. FPGA-ROS~\cite{Podlubne2020} allows for having multiple hardware accelerators per FPGA. They are connected through a streaming network that provides connectivity to other ROS nodes using a central Ethernet gateway. The communication is supervised by a central managing instance.

\subsection{ReconROS and fpgaDDS}
\label{sec:BackgroundRelatedWork:ReconROS}

ReconROS~\cite{lienen2021design} is a framework for hardware acceleration of ROS 2 applications. It combines the reconfigurable hardware operating system ReconOS~\cite{Luebbers_Platzner_2009} with ROS 2 and allows for implementing complete ROS 2 nodes in hardware. Due to ReconROS API, both software-mapped and hardware-mapped ROS 2 nodes use the same consistent programming model.

\begin{figure}[h]
    \centering
    \includegraphics[width=0.75\linewidth]{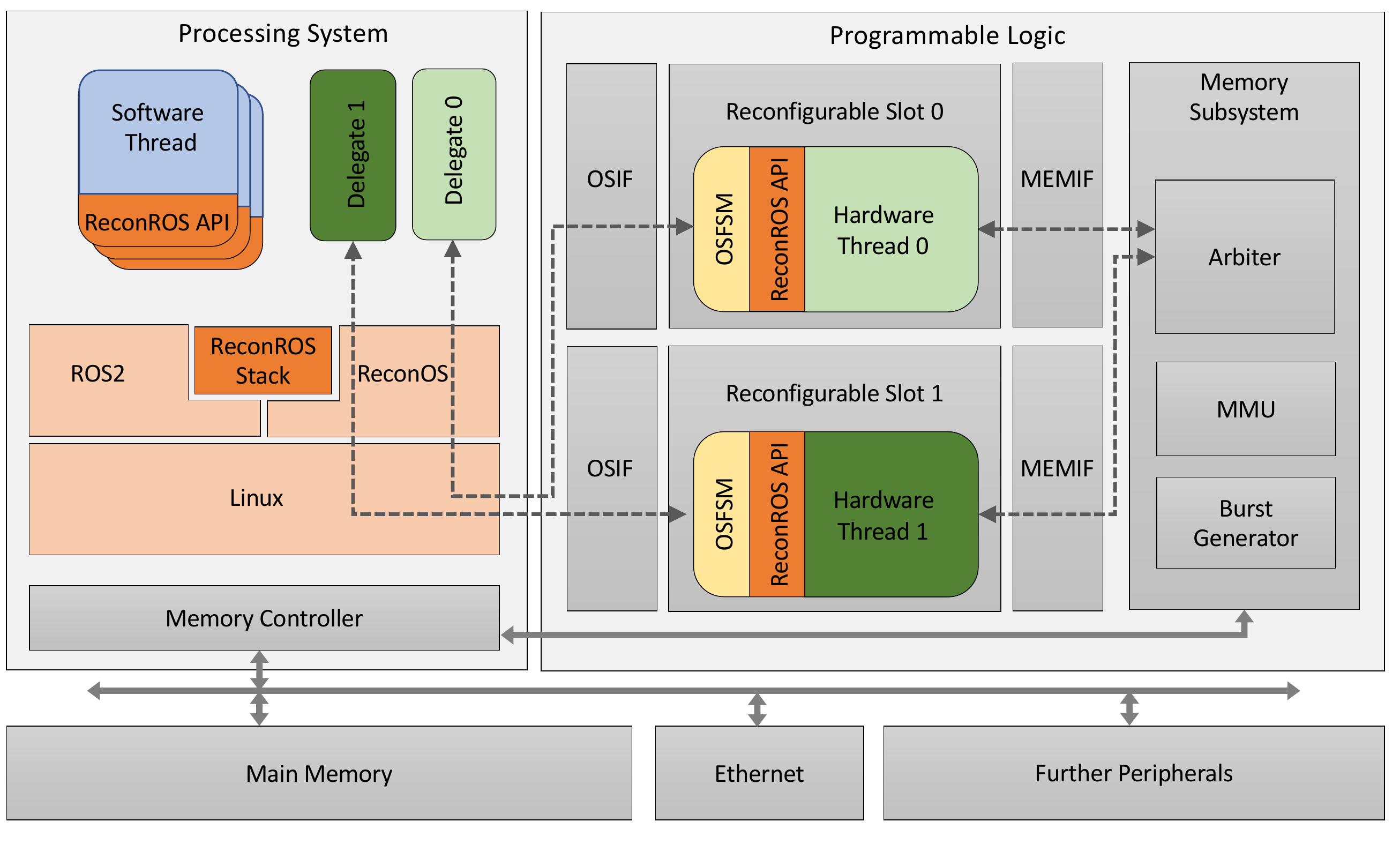}
    \caption{ReconROS architecture (from~\cite{lienen2021design}) }
    \label{fig:background:reconros_architecture}
\end{figure}

Figure~\ref{fig:background:reconros_architecture} shows the architecture of ReconOS comprising two hardware threads and three software threads. Each of the hardware threads is mapped to a reconfigurable slot providing two interfaces for communication: First, the MEMIF interface enables memory accesses in virtual address space. A memory management unit part of the ReconOS memory subsystem translates to physical addresses. Second, the OSIF interface connects to the hardware thread's delegate thread. The delegate thread is a lightweight software thread interacting with the operating system and ROS 2 software layers on behalf of the hardware thread. 
ReconROS extends ReconOS by the ReconROS API and the ReconROS stack. The ReconROS stack includes ROS-related objects, e.g., ROS nodes, subscribers, and messages. Due to the ReconROS API in hardware and software, ReconROS provides a consistent programming model for both software and hardware ROS 2 nodes interacting with the ReconROS stack.

When a hardware-mapped ROS 2 node subscribes to a topic and wants to read a message, the following steps are performed: The hardware thread implementing the ROS 2 node sends a request through its OSIF interface to its delegate thread. The delegate thread receives the request and calls a blocking read operation at the ROS 2 layer. After receiving the message, the call unblocks, and the delegate sends the pointer to the received message via the OSIF interface back to the hardware thread. Then, the hardware thread can access the message in the main memory through its MEMIF interface, typically copying it into an FPGA-internal buffer. Since all hardware threads and, thus, all hardware-mapped ROS 2 nodes share the ReconROS memory subsystem, the available memory bandwidth is divided by the number of nodes that simultaneously access messages. 

fpgaDDS~\cite{lienen2023fpgadds} is an extension to ReconROS that aims at accelerating the communication between hardware-mapped nodes. fpgaDDS provides a static intra-FPGA data distribution service (DDS) automatically generated during design time. fpgaDDS 
leverages separate streaming networks between hardware threads and features so-called hardware-mapped topics (HMT), in contrast to the standard software-mapped topics (SMT) that use buffer implementations in main memory.
\section{Gateway Design}
\label{sec:Gateway}

We design a gateway to close the gap between software-mapped topics (SMT) and hardware-mapped topics (HMT). All hardware-mapped ROS 2 nodes that have to communicate with software-mapped nodes share one MEMIF interface to the main memory and, thus, the memory bandwidth. Gateways aim at reducing the number of data transfers per message in such cases to one.

\begin{figure}[h]
  \centering
  \includegraphics[width=0.75\linewidth]{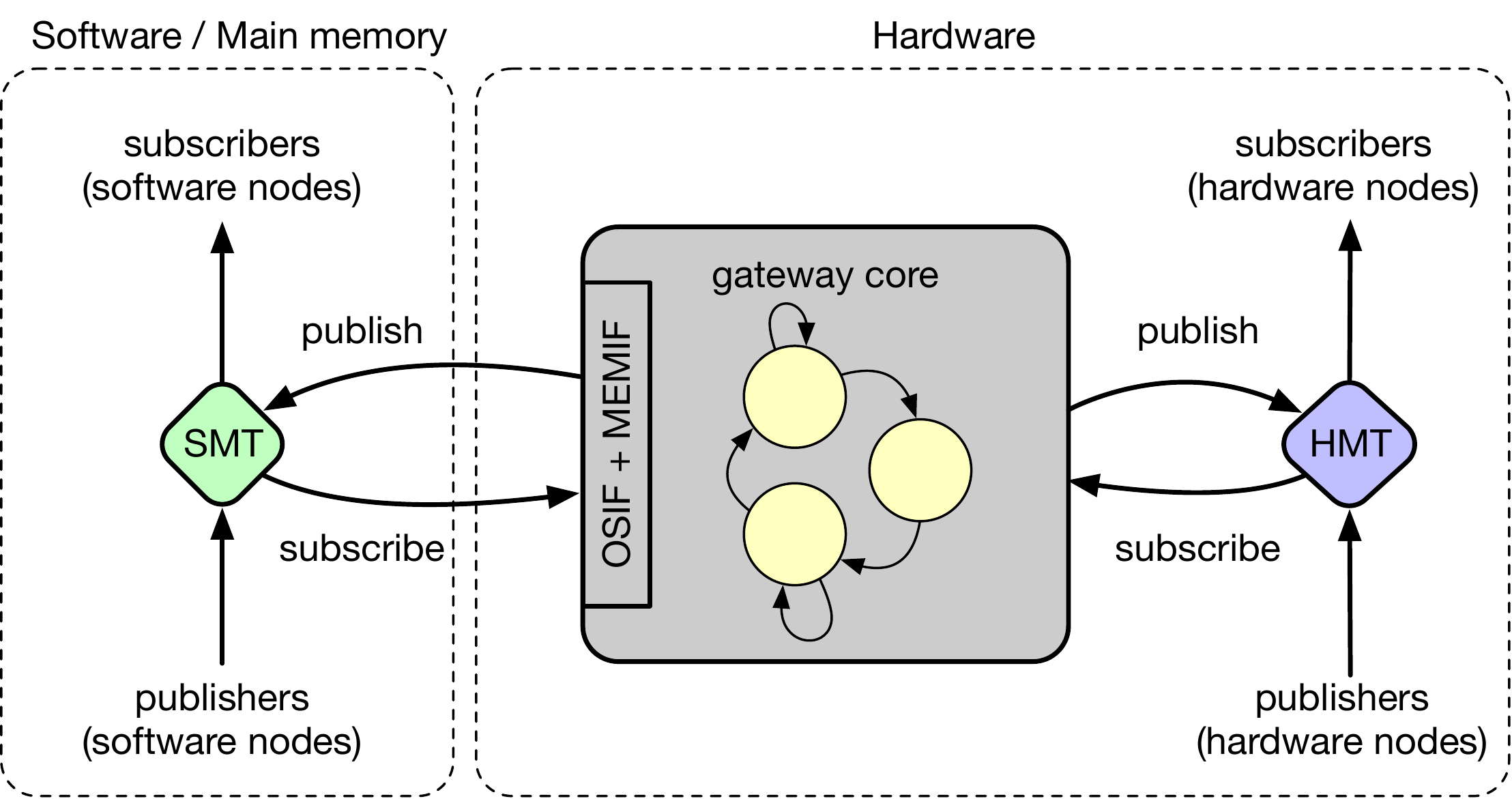}
  \caption{Gateway Architecture}
  \label{fig:gateway:architecture}
\end{figure}

Figure~\ref{fig:gateway:architecture} sketches the architecture of the gateway. A gateway comprises three main components, a software-mapped topic, a hardware-mapped topic, and the gateway core. Internally, the gateway core is implemented similarly to a node mapped to hardware and establishes publish-subscribe channels to both the software-mapped and hardware-mapped topics. Other hardware-mapped ROS 2 nodes publishing or subscribing to the topic connect to the gateway's HMT, and other software-mapped ROS 2 nodes to its SMT. Since the gateway core synchronizes the SMT and the HMT, only one data transfer to or from the main memory, i.e., between the software and hardware domains, is required per message, significantly reducing the required memory bandwidth.

\begin{figure}[h]
  \centering
  \includegraphics[width=0.85\linewidth]{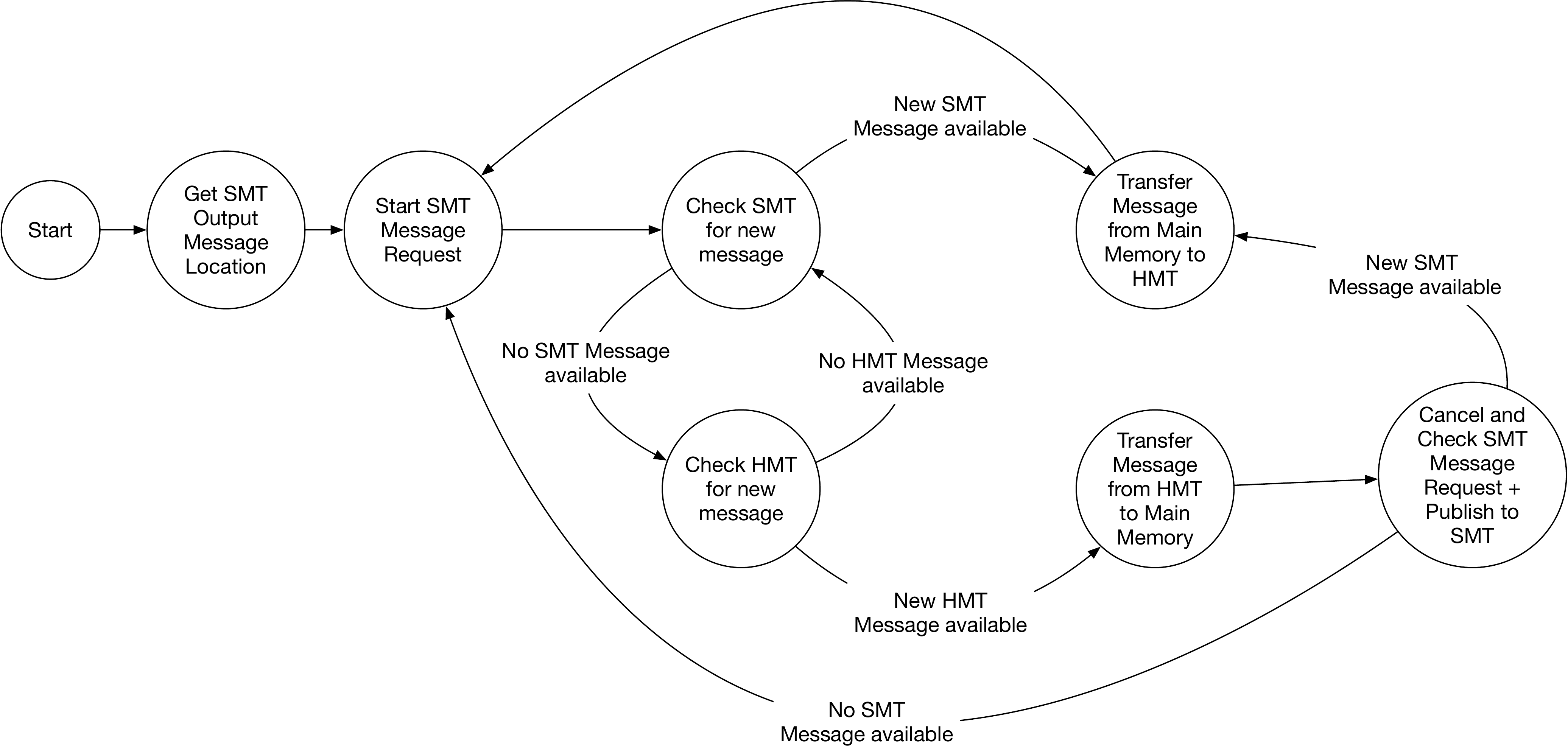}
  \caption{Runtime behavior of the gateway core}
  \label{fig:gateway:fsm}
\end{figure}

The finite state machine in Figure~\ref{fig:gateway:fsm} presents the runtime behavior of the gateway core: At startup, the gateway core receives the location of the output message for its SMT from its delegate thread and sends a request to it for a new SMT message. Upon that, the delegate thread blocks and waits for new messages. After receiving a new message, it responds to the gateway core through the OSIF interface. The gateway core polls both its OSIF for a response from its delegate thread and its HMT for a new message from a hardware-mapped node. In the first case, the message is transferred to the HMT. In the second case, the gateway core transfers the message to the main memory and then cancels the message request to the delegate thread before it publishes the data to the SMT. Since a new message on the SMT could potentially be received between request and cancel, the cancel process may respond with a pointer to a new message. This message is transferred to the HMT before the gateway starts a new request to its delegate. To avoid loops in the gateway core, for example, by messages that are received from an SMT, republished in the HMT, and then, are received again by the gateway's subscriber on the HMT side, our gateway core implementation includes message filters in both its subscribers on the SMT and HMT sides. These filters check and discard messages for their publisher IDs if the publisher source and destination IDs match.

\section{Performance Measurements}
\label{sec:Measurements}

After describing the architecture and functionality of the gateway, in this section, we evaluate the gateway performance through a synthetic setup. The measurements aim to characterize situations where a gateway should be preferred over an SMT and, thus, develop recommendations for the communication mapping step. 

All ROS 2 hardware nodes and the gateways used for performance measurements have been implemented in C/C++ and synthesized to a hardware description language (HDL) format with the high-level synthesis tool Vivado Vitis 2021.2. The HDL codes and the ReconROS infrastructure were then synthesized to an FPGA bitstream. Software nodes, including the software part of ReconROS, have been compiled using gcc. We have leveraged the ZCU104 evaluation board comprising an UltraScale+ MPSoC FPGA running Ubuntu Linux 20.04, ReconROS, and ROS 2 galactic for the measurements.

\begin{figure}[!h]
  \centering
  \includegraphics[width=0.85\linewidth]{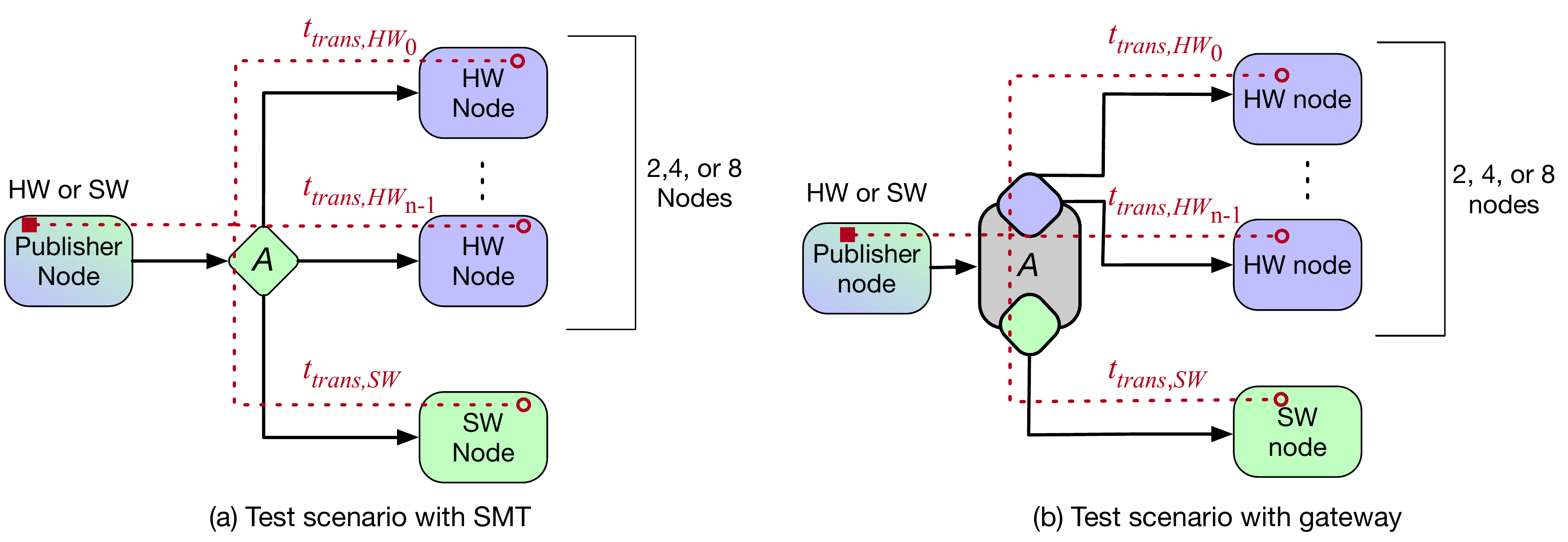}
  \caption{Test setup comprising one publishing node (hw or sw), $2$, $4$, or $8$ subscribing hardware nodes and one subscribing software node}
  \label{fig:evaluation:setup}
\end{figure}

Figure~\ref{fig:evaluation:setup} sketches the experimental setup. In Figure~\ref{fig:evaluation:setup}(a), a publishing node, that is either a software-mapped or a hardware-mapped node, generates messages with random data and publishes it to a topic $A$. The message type is Image from the sensor message package of ROS 2 with image data of $10\:kB$, $100\:kB$, $1\:MB$, and $10\:MB$. A set of receiving nodes subscribe to topic $A$ and copy the received message into their local memory. The subscribing node set comprises one software node and $2$, $4$, or $8$ hardware nodes. Figure~\ref{fig:evaluation:setup}(b) shows the same setup but with a gateway $A$ instead of an SMT $A$. The overall 12 experiments have been repeated $500$ times, and the mean values of the measured transmission times are reported.

\begin{figure}[!h]
  \centering
  \includegraphics[width=1\linewidth]{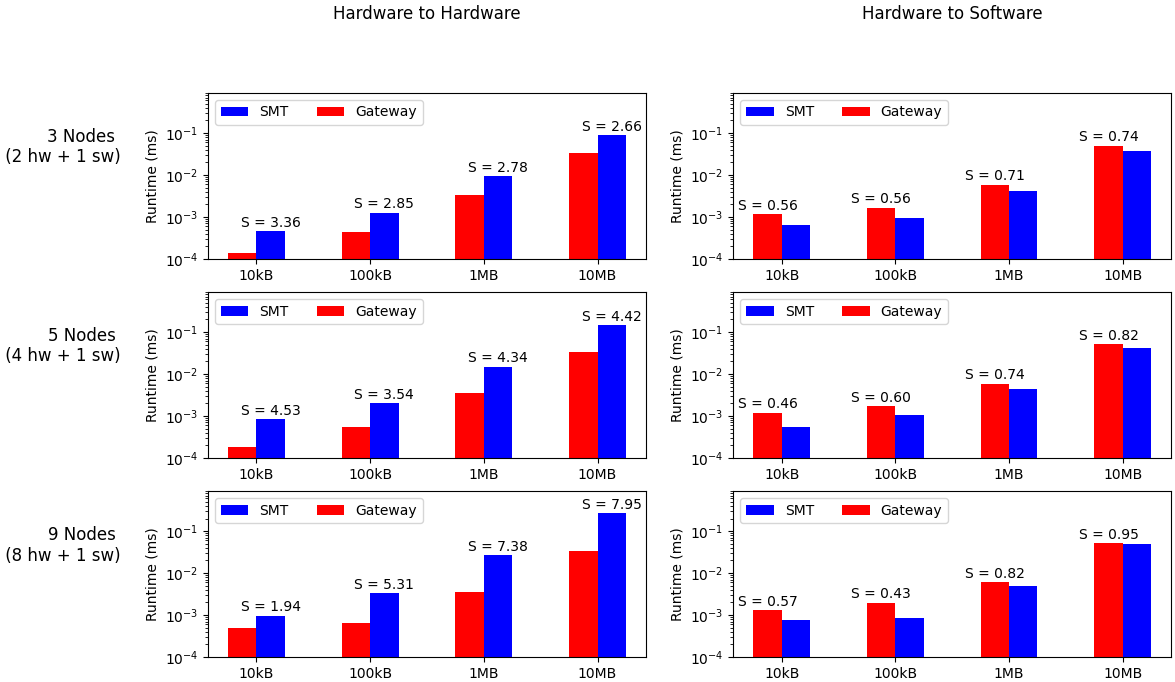}
  \caption{Measured maximum transfer times for hardware-to-hardware and hardware-to-software communication and the resulting speedups}
  \label{fig:evaluation:transfertimes_hw}
\end{figure}

Figure~\ref{fig:evaluation:transfertimes_hw} shows the results for the hardware publisher node. The left column of the figure reports the maximum transmission times to any of the subscribing hardware nodes: $t_{trans,HW} = \max_{0 \leq i < n} \{t_{trans, HW_i}\}$. The results show significant speedups for the gateway compared to an SMT as soon as we have more than one hardware-mapped subscribing node. The range of speedups depends on the message size and, for the example of 8 hardware nodes, ranges from $1.94$ for small messages to $7.95$ for larger messages.

The right column of Figure~\ref{fig:evaluation:transfertimes_hw} presents the transmission times for the publishing hardware node to the subscribing software node. Due to its internal design, the gateway introduces overhead for the hardware-to-software transmission. This overhead is significant for smaller message sizes and results in larger transmission times up to a factor of approximately $2$. However, for larger message sizes and an increasing number of hardware nodes, the overhead reduces and the speedup approaches $1$. 

\begin{figure}[!h]
  \centering
  \includegraphics[width=1\linewidth]{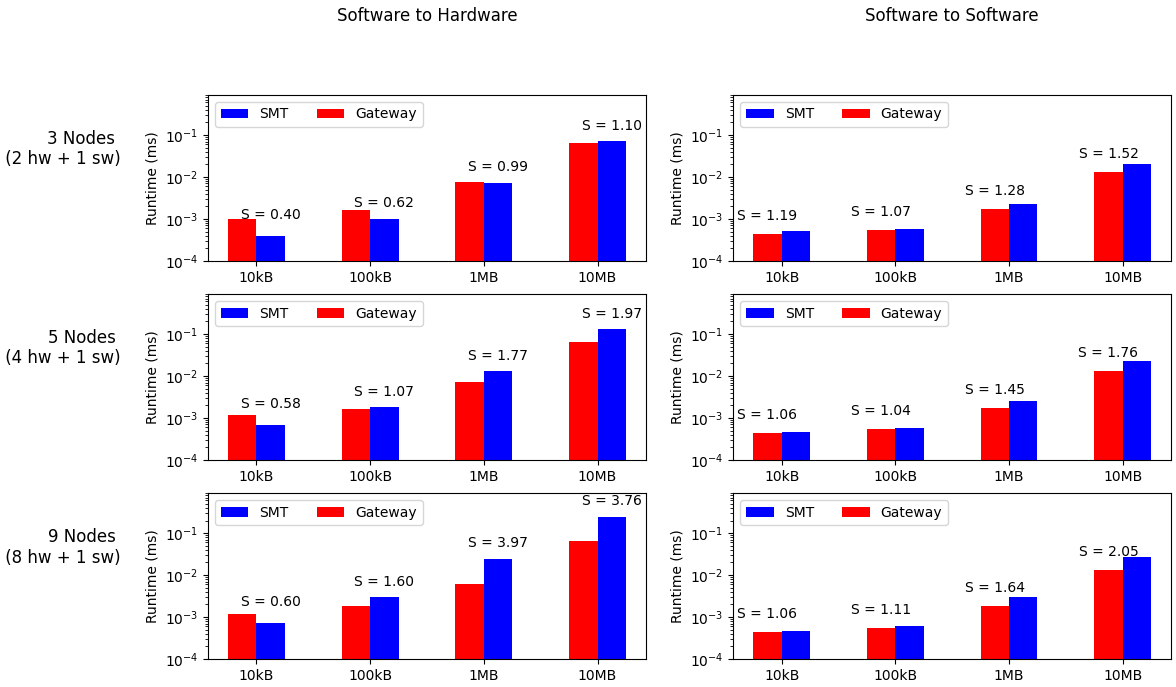}
  \caption{Measured maximum transfer times for software-to-hardware and software-to-software communication and the resulting speedups}
  \label{fig:evaluation:transfertimes_sw}
\end{figure}

Figure~\ref{fig:evaluation:transfertimes_sw} shows the results for the software publisher node. 
Again, the left column of the figure displays the maximum transmission times to any of the subscribing hardware nodes. For smaller message sizes, the gateway overhead leads to speedups below $1$. Still, for larger message sizes, we achieve speedups up to $3.79$, below those achieved for the hardware publisher node. 

The right column of Figure~\ref{fig:evaluation:transfertimes_sw} presents the transmission times for the publishing software node to the subscribing software node. Here, we achieve speedups for the gateway for all measured message sizes and numbers of hardware nodes. 
The range of speedups for software-to-software communication is from $1.06$ to $2.05$. 

In conclusion, publish-subscribe communication with larger message sizes and multiple involved hardware-mapped subscriber nodes significantly benefits from using gateways since the required memory bandwidth on the MEMIF is minimized in such situations. 
The benefits become less significant for smaller message sizes and fewer hardware-mapped nodes. 

\section{Mapping ROS Applications}
\label{sec:DesignMethodology}

A ROS computation graph can be formally expressed as directed graph $\mathcal{G}=(\mathcal{V},\mathcal{E})$, where the set of graph nodes $\mathcal{V}$ comprises both ROS nodes and ROS topics, i.e., $\mathcal{V}=(\mathcal{N},\mathcal{T})$, and the set of graph edges $\mathcal{E}$ splits into edges indicating a publish function and edges denoting a subscribe function, i.e., $\mathcal{E}=(\mathcal{E}_{pub},\mathcal{E}_{sub})$ with $\mathcal{E}_{pub} = \{ (x,y) \, | \, x \in \mathcal{N}, y \in \mathcal{T} \}$ and $\mathcal{E}_{sub} = \{ (x,y) \, | \, x \in \mathcal{T}, y \in \mathcal{N} \} $. Additionally, we define for each topic $t \in \mathcal{T}$ the set of publishing and subscribing ROS nodes as $\mathcal{E}_{pub}^t = \{ (x,y) \in \mathcal{E}_{pub} \, | \, y = t\}$ and $\mathcal{E}_{sub}^t = \{ (x,y) \in \mathcal{E}_{sub} \, | \, x = t\}$.

\begin{figure}[!h]
	\center
    \includegraphics[width=1.0\linewidth]{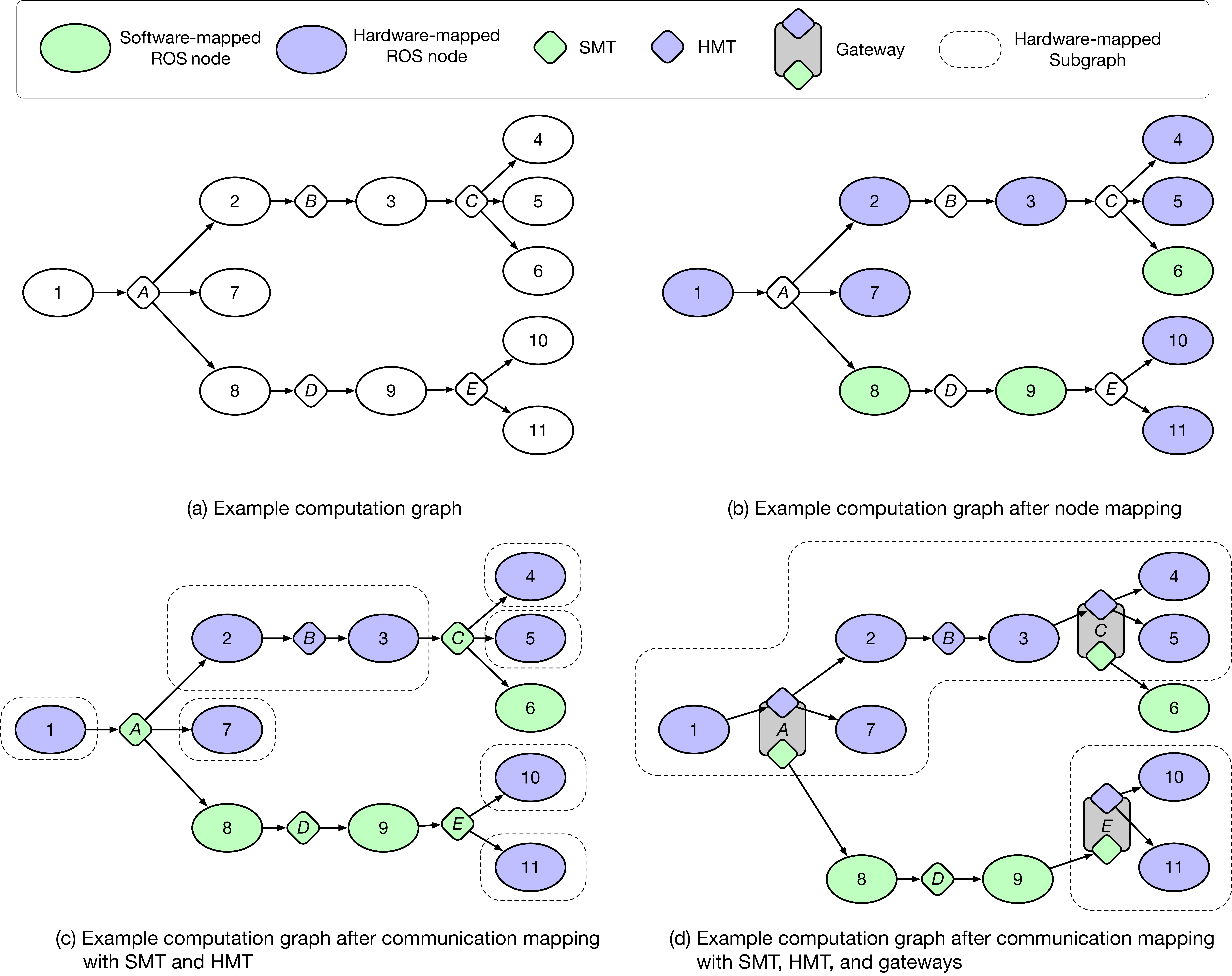}
    \caption{Example for node and communication mapping}
    \label{fig:computationgraph}
\end{figure}

In our design flow, mapping a ROS application to a cSoC comprises two subsequent steps: (i) node mapping and (ii) communication mapping. Figure~\ref{fig:computationgraph} presents these steps on an exemplary computation graph. 

Starting from the original computation graph in Figure~\ref{fig:computationgraph}(a), the node mapping step assigns each node $n \in \mathcal{N}$ to either a hardware or a software implementation. We denote the set of nodes mapped to hardware as $\mathcal{N}_{HW}$ and the set of nodes mapped to software as $\mathcal{N}_{SW}$ and, obviously, the node mapping must satisfy $\mathcal{N} = \mathcal{N}_{HW} \cup \mathcal{N}_{SW}$. We currently envision that the developer decides whether to map a specific node to software or hardware. This decision will depend on, e.g., whether an accelerated or energy-efficient implementation is desirable and available for the node and whether there is logic capacity left in the FPGA. Figure~\ref{fig:computationgraph}(b) shows the result of an exemplary node mapping phase, where $\mathcal{N}_{HW} = \{1, 2, 3, 4, 5, 7, 10, 11\}$ and $\mathcal{N}_{SW} = \{6, 8, 9\}$. 

The second step is communication mapping, where we assign each topic $t \in \mathcal{T}$ an implementation in software, hardware, or as a gateway. We denote the set of topics mapped to software as $\mathcal{T}_{SW}$, the set of topics mapped to hardware as $\mathcal{T}_{HW}$, and the set of topics mapped to a gateway as $\mathcal{T}_{GW}$. Obviously, the communication mapping must satisfy $\mathcal{T} = \mathcal{T}_{SW} \cup \mathcal{T}_{HW} \cup \mathcal{T}_{GW}$. 

The default mapping for topics is to the software since software-mapped topics are the most flexible and can connect any number of software and hardware-mapped nodes. However, if all topics of our running example were actually mapped to software, there are overall ten edges to and from hardware-mapped nodes, three of them are in $\mathcal{E}_{pub}$, and seven are in $\mathcal{E}_{sub}$. Each of these edges will lead to data transfer buffered in the cSoC's main memory to or from the configurable logic. When all hardware-mapped nodes execute in parallel, which is a desired scenario and the motivation for hardware acceleration, all these transfers will have to share the available memory bandwidth of ReconROS' MEMIF.

Therefore, we optimize the communication mapping by identifying three cases. First, we search for topics in the node-mapped computation graph for which all the publishers and subscribers are mapped to software. For such topics, the standard software implementation is selected. Formally, we check for each topic $t \in \mathcal{T}$ the following condition: $(\forall (x,t) \in \mathcal{E}_{pub}^t: x \in \mathcal{N}_{SW}) \wedge (\forall (t,y) \in \mathcal{E}_{sub}^t: y \in \mathcal{N}_{SW})$. In our example, the condition only holds for topic $D$.

Second, we identify topics in the node-mapped computation graph for which all the publishers and subscribers are mapped to hardware. Such topics will then be mapped to hardware as well, realized with dedicated hardware components of the fpgaDDS layer~\cite{lienen2023fpgadds}. Compared to software topics, hardware topics provide much higher communication bandwidth and reduced latency. Formally, we check for each topic $t \in \mathcal{T}$ the following condition: $(\forall (x,t) \in \mathcal{E}_{pub}^t: x \in \mathcal{N}_{HW}) \wedge (\forall (t,y) \in \mathcal{E}_{sub}^t: y \in \mathcal{N}_{HW})$. In our example, the condition only holds for topic $B$.

The remaining topics, for which neither of the above conditions holds, connect software and hardware-mapped nodes. Figure~\ref{fig:computationgraph}(c) shows the resulting computation graph if such topics are realized in software. The resulting communication mapping is $\mathcal{T}_{SW} = \{A,C,D,E\}$ and $\mathcal{T}_{HW} = \{B\}$. The figure further indicates subgraphs that are mapped to hardware with dashed lines, and it can be seen that the number of edges crossing the software-hardware boundary is reduced from 10 to eight, two of them are in $\mathcal{E}_{pub}$, and six are in $\mathcal{E}_{sub}$. 

However, an inspection of the topics $A, C$, and $E$ reveals that they connect to more than one hardware-mapped node of subscribers. Thus, according to the characterization discussed in Section~\ref{sec:Measurements}, it is beneficial, at least for larger message sizes, to implement these topics as gateways. Figure~\ref{fig:computationgraph}(d) shows the resulting computation graph mapping with 
$\mathcal{T}_{SW} = \{D\}$, $\mathcal{T}_{HW} = \{B\}$, and $\mathcal{T}_{GW} = \{A, C, E\}$. Again, the figure indicates subgraphs that are mapped to hardware with dashed lines, and the number of edges crossing the software-hardware boundary is finally reduced to 3, one from gateway $A$ to the software-mapped node 8, one from gateway $C$ to the software-mapped node 6, and the last one from the software-mapped node 9 to the gateway $E$.

\section{Design Example Application}
\label{sec:Evaluation}

This section reports on evaluating a gateway used in a simulated real-world scenario. In this scenario, we elaborate on an extended implementation of the autonomous driving architecture from \cite{lienen2023fpgadds}. In this work, the speedup ($2.48$) due to the usage of hardware-mapped topics compared to standard ROS 2 communication has already been shown. The simulated environment for the autonomous driving example is realized in Gazebo under Ubuntu 20.04 with ROS 2 galactic running on a desktop PC with Intel Core i5-8000 CPU connected via Gigabit Ethernet to the evaluation board. The remaining setup is inherited from Section \ref{sec:Measurements}.

\begin{figure}[!h]
  \centering
  \includegraphics[width=1\linewidth]{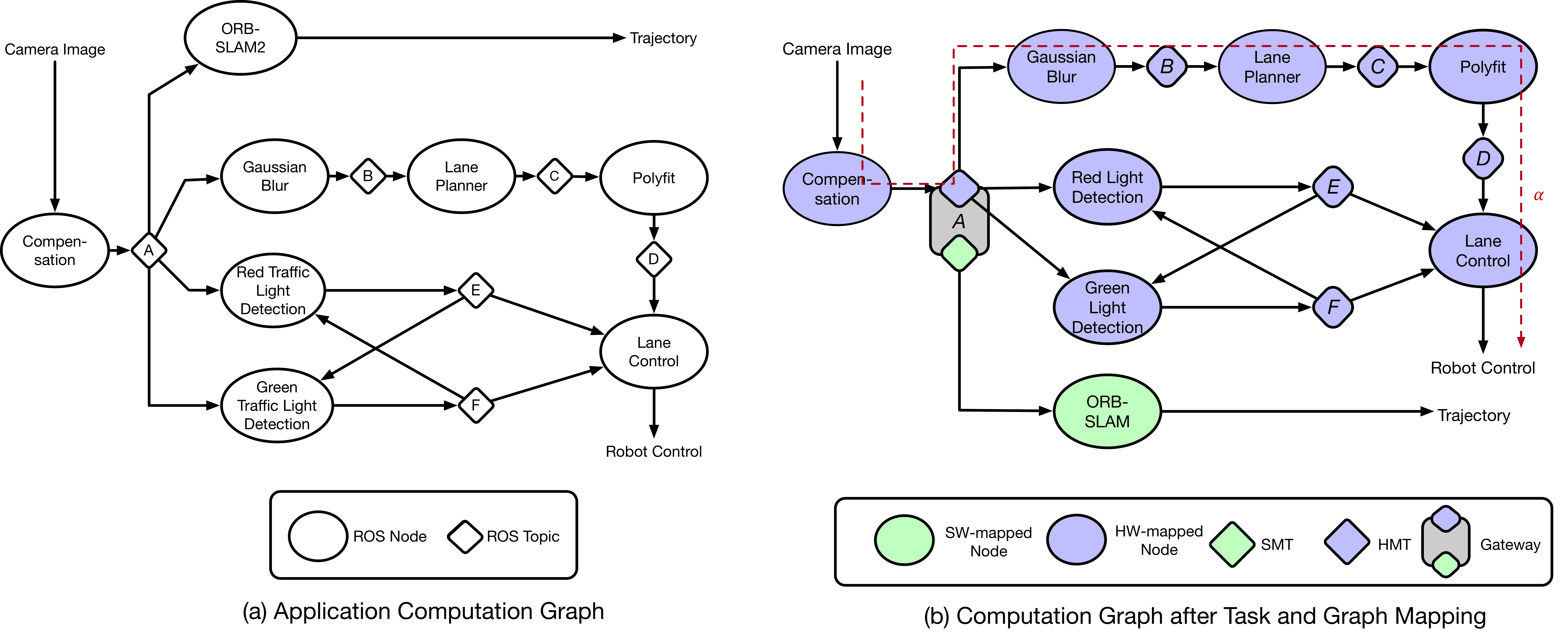}
  \caption{Example Design}
  \label{fig:evaluation:exampledesign}
\end{figure}

Compared to \cite{lienen2023fpgadds}, we modified the driving example in three steps: First, we optimized the driving behavior of the autonomous robot with substitution of the {\itshape Image Projection} and {\itshape Lane Following} node by the {\itshape Lane Planner} and {\itshape Polyfit} node (cf. Figure \ref{fig:evaluation:exampledesign}). {\itshape Lane Planner} computes color space conversation to the HSV color space. The node generates a mask for yellow and white pixels from the transformed image by checking pixel-wise for specific ranges. After combining both masks, the perspective of the combined mask is transformed into a bird's view. The resulting image is published on a hardware-mapped topic. The {\itshape Polyfit} node subscribes to that HMT and generates an approximation of the lane out of it, solving a least-squares problem. Second, we extended the computation graph by a node implementing the popular ORB-SLAM 3 algorithm \cite{orbslam3} for creating a map of the environment. Without using gateways for hardware-mapped topics, the subscription of the output of the {\itshape Image Compensation} node would result in falling back to standard ROS 2 communication for the topic $A$ and, therefore, potentially slow down the application due to memory transfer overheads. Third, we changed the simulated environment of the robot to an urban environment by adding buildings. This allows the ORB-SLAM3 algorithm to find remarkable key points for creating a map.

The remaining nodes are implemented as follows: first, the {\itshape Compensation} node calculates a histogram of the image and removes outlier pixels of the input image. The {\itshape Gaussian Blur} node performs a Gaussian low-pass filter on the image. The {\itshape Lane Control} node subscribes to the output of the {\itshape Polyfit} node and calculates drive commands out of it. The {\itshape Green Traffic Light Detection} and {\itshape Red Traffic Light Detection} nodes both subscribe to the corrected image from the {\itshape Image Compensation} node. Both scan the input image for red or green traffic lights, respectively. If successful, the nodes publish to a separate output topic, which is received by the other node and by the {\itshape Lane Control} node. If a red traffic light is detected, the robot stops, and the green traffic light detection is activated. If the traffic light is green, the detection for red traffic lights is activated, and the car starts driving again.

We implemented the computation graph in two versions: The first version uses a hardware-mapped topic gateway to map topic $A$ into the hardware domain. For the second version, we use standard ROS 2 communication. For both versions, we have measured the execution time of the node chain $\alpha$ (red dotted line in Figure \ref{fig:evaluation:exampledesign}). For the implementation using standard ROS 2 communication for topic $A$, the mean execution time for the node chain is $28.264\:ms$ with a standard deviation of $3.21\:ms$. For the implementation leveraging the proposed gateway, we measured a mean execution time of $20.270\:ms$ and a standard deviation of $0.239\:ms$. We achieved an average speedup of $1.4 \times $ and a one-order of magnitude reduced standard deviation.
\section{Conclusion and Future Work}
\label{sec:Conclusion}

This paper presents gateways, a novel communication infrastructure component closing the gap between standard ROS 2 communication and hardware-mapped topics. Gateways reduce data transfers between hardware and software, leading to more efficient utilization of configurable system-on-chip. Building on gateways, we present a methodology for mapping inter-node communication in hardware-accelerated ROS 2 applications with ReconROS and fpgaDDS. The method aims to find subgraphs of the ROS 2 computation graph for mapping completely to hardware benefiting from higher available bandwidth and lower latencies. 

For future work, we plan to extend the methodology for task mapping to complete the mapping of ROS 2 computation graphs to system-on-chip architectures. However, the proposed methodology does not consider applications leveraging dynamically mapped hardware nodes, which offers additional research potential for more efficient hardware utilization.

\bibliographystyle{ACM-Reference-Format}
\bibliography{lienen23_icrai}

\end{document}